\pdfoutput=1

\documentclass[11pt]{article}

\usepackage[final]{acl}

\usepackage{times}
\usepackage{latexsym}
\usepackage[linesnumbered,ruled,vlined]{algorithm2e}
\usepackage{bm}
\usepackage{algorithmicx}
\usepackage{amsmath}
\usepackage{amssymb}
\usepackage{booktabs}
\usepackage{multirow}
\usepackage{natbib}
\usepackage{adjustbox}
\usepackage{tcolorbox}
\usepackage{subfigure}

\usepackage[T1]{fontenc}

\usepackage[utf8]{inputenc}

\usepackage{microtype}

\usepackage{inconsolata}

\usepackage{graphicx}

\title{GuiLoMo: Allocating Expert Number and Rank for \underline{Lo}RA-\underline{Mo}E via Bilevel Optimization with \underline{Gui}dedSelection Vectors}

\makeatletter
\def\@fnsymbol#1{}
\makeatother

\author{
\textbf{Hengyuan Zhang\textsuperscript{1 *}},
 \textbf{Xinrong Chen\textsuperscript{2 *}}\thanks{*\ Equal contribution. Order set by random dice roll.},
 \textbf{Yingmin Qiu\textsuperscript{3}},
 \textbf{Xiao Liang\textsuperscript{4}},
 \textbf{Ziyue Li\textsuperscript{5}},\\
 \textbf{Guanyu Wang\textsuperscript{2}},
 \textbf{Weiping Li\textsuperscript{2}},
 \textbf{Tong Mo\textsuperscript{2 \dag}}\thanks{\dag\ Corresponding author.},
 \textbf{Hayden Kwok-Hay So\textsuperscript{1}},
 \textbf{Ngai Wong\textsuperscript{1 \dag}}
\\
 \textsuperscript{1}The University of Hong Kong, 
 \textsuperscript{2}Peking University, \\
 \textsuperscript{3}Beijing University of Posts and Telecommunications, 
 \textsuperscript{4}University of California, Los Angeles, \\
 \textsuperscript{5}National Science Library, Chinese Academy of Sciences
 \\
\texttt{hengyuan.zhang88@gmail.com \ \ chenxinrong23@stu.pku.edu.cn}  
}

\begin{document}
\maketitle
\begin{abstract}
Parameter-efficient fine-tuning (PEFT) methods, particularly Low-Rank Adaptation (LoRA), offer an efficient way to adapt large language models with reduced computational costs. However, their performance is limited by the small number of trainable parameters. Recent work combines LoRA with the Mixture-of-Experts (MoE), i.e., LoRA-MoE, to enhance capacity, but two limitations remain in hindering the full exploitation of its potential: 1) the influence of downstream tasks when assigning expert numbers, and 2) the uniform rank assignment across all LoRA experts, which restricts representational diversity.
To mitigate these gaps, we propose GuiLoMo, a fine-grained layer-wise expert numbers and ranks allocation strategy with GuidedSelection Vectors (GSVs). GSVs are learned via a prior bilevel optimization process to capture both model- and task-specific needs, and are then used to allocate optimal expert numbers and ranks.
Experiments on three backbone models across diverse benchmarks show that GuiLoMo consistently achieves superior or comparable performance to all baselines. Further analysis offers key insights into how expert numbers and ranks vary across layers and tasks, highlighting the benefits of adaptive expert configuration. Our code is available at \url{https://github.com/Liar406/Gui-LoMo.git}.
\end{abstract}

\section{Introduction}
\begin{table}[!t]
\centering
\setlength\tabcolsep{5pt}
\fontsize{9.5}{13}\selectfont 
\begin{tabular}{lccc}
\toprule[1.5pt]
                & MoLA & AlphaLoRA & \begin{tabular}[c]{@{}c@{}}GuiLoMo\\ (Ours)\end{tabular} \\ \hline \addlinespace[2pt]
Model Specific & \adjustbox{valign=b}{\includegraphics[width=0.03\textwidth]{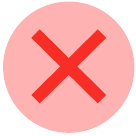}}    & \adjustbox{valign=b}{\includegraphics{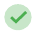}}         & \adjustbox{valign=b}{\includegraphics{correct.pdf}}                                                        \\
Task Specific  & \adjustbox{valign=b}{\includegraphics[width=0.03\textwidth]{wrong.pdf}}    & \adjustbox{valign=b}{\includegraphics[width=0.03\textwidth]{wrong.pdf}}         & \adjustbox{valign=b}{\includegraphics{correct.pdf}}                                                        \\
Expert Number  & \adjustbox{valign=b}{\includegraphics{correct.pdf}}    & \adjustbox{valign=b}{\includegraphics{correct.pdf}}         & \adjustbox{valign=b}{\includegraphics{correct.pdf}}                                                        \\
Expert Rank    & \adjustbox{valign=b}{\includegraphics[width=0.03\textwidth]{wrong.pdf}}    & \adjustbox{valign=b}{\includegraphics[width=0.03\textwidth]{wrong.pdf}}         & \adjustbox{valign=b}{\includegraphics{correct.pdf}}                                                        \\ \bottomrule[1.5pt]
\end{tabular}
\vspace{-0.2cm}
\caption{Compared to existing methods, our proposed GuiLoMo strategy can allocate the optimal expert numbers and ranks within LoRA-MoE, tailored to specific models and tasks.}
\label{tab:intro}
\vspace{-0.5cm}
\end{table}

Although large language models (LLMs) have demonstrated remarkable performance across a wide range of general tasks~\citep{jiang2023mistral7b,chowdhery2023palm,jian2023expedited,touvron2023llama,xiong2023dq}, they still fall short in certain tasks or domains, such as reasoning~\citep{tong2024can,srivastava2024mathdivide,yu2025chain,li2025system,liang2025sws}, multilingualism~\citep{huang2023not,gurgurov2024multilingual,zhang2024shifcon}, and text generation in specialized contexts~\citep{biancotti2024chat,zhang2024balancing,yang2024llm,yang2025quantifying,li-etal-2024-simulating,li2024dalk,wang2024bpo,chang2025treereview}.
To enhance the performance of LLMs in these challenging areas, a common practice is fine-tuning. However, with the growing size of current LLMs, full fine-tuning faces significant challenges in terms of computational efficiency and memory consumption. To mitigate these issues, parameter-efficient fine-tuning (PEFT) methods have gained considerable attention~\citep{houlsby2019parameter,li-liang-2021-prefix,lester-etal-2021-power,hu2022lora, liu-etal-2022-p, zhang2023adaptive, yang2025mtl}. Among these methods, Low-Rank Adaptation (LoRA)~\citep{hu2022lora} is regarded as one of the most efficient approaches. Nonetheless, its performance remains constrained due to the relatively small number of trainable parameters~\citep{xu2023parameter}.
Recent studies suggest that combining LoRA with the Mixture-of-Experts (MoE) paradigm, referred to as LoRA-MoE, by incorporating multiple LoRA modules, offers a promising solution to this limitation~\citep{wu2024mixture,gao2024higher,qing-etal-2024-alphalora,dou-etal-2024-loramoe,liu2023moelora,luo2024moelora}.

However, fully exploiting the potential of LoRA-MoE remains an open research question. First, \citet{gao2024higher} considered that uniformly allocating the number of experts across all layers is suboptimal, as different layers play distinct roles in the model. Over-allocating experts to certain layers can lead to redundancy and degraded performance. To address this, they proposed a group-wise expert allocation strategy (MoLA), which divides all layers into four groups and assigns varying numbers of experts to each group, ensuring that layers within the same group share the same number of experts. Building on this, \citet{qing-etal-2024-alphalora} introduced a layer-wise allocation strategy (AlphaLoRA), which theoretically determines the expert numbers for each layer based on its training quality.

Despite these advancements, two critical limitations remain, as shown in Table~\ref{tab:intro}:
1) These methods determine the expert number without considering the downstream task. This is problematic, as different tasks may have varying levels of complexity and specific needs, which should influence the optimal expert configuration (as supported by experiments in Appendix~\ref{sec:Diverse_Downstream}); 2) These methods also overlook the intrinsic rank of LoRA experts, typically assigning the same rank to all LoRA experts. This uniformity leads to equivalent representational capacities across experts, causing them to capture similar information. Thus, LoRA-MoE struggles to handle diverse and complex inputs.

To address these limitations, we propose GuiLoMo, a fine-grained strategy for jointly allocating layer-wise expert numbers and ranks in LoRA-MoE based on bilevel optimization with GuidedSelection vectors. GuiLoMo operates in two steps:
1) Obtaining GuidedSelection Vectors (GSVs): Through an initial optimization, GSVs are learned to guide LoRA-MoE in selecting the optimal expert numbers and ranks tailored to both the model backbone and the downstream task;
2) Allocating Expert Numbers and Ranks: After the prior optimization, the optimized GSVs are used to allocate expert numbers and ranks for LoRA-MoE, followed by the final training phase.

To summarize, our contributions are as follows:

\vspace{0.2em} 
\quad \textbf{1)} To further unlock the potential of LoRA-MoE, we propose GuiLoMo, a fine-grained layer-wise expert numbers and ranks allocation strategy based on proposed GuidedSelection Vectors.

\vspace{0.2em}
\quad \textbf{2)} We conduct extensive experiments on a wide range of tasks, including natural language understanding, question answering, and mathematical reasoning, demonstrating the effectiveness of GuiLoMo. For instance, GuiLoMo achieves an average $2.61\%$ improvement on mathmatical reasoning tasks with $\text{LLaMA-2}_\text{7B}$. Further analysis confirms the effectiveness of GuidedSelection vectors in selecting optimal expert numbers and ranks.

\vspace{0.2em}
\quad \textbf{3)} We provide valuable insights into the relationship between expert numbers, ranks, and their assigned layers. 
For example, we observe that multi-head attention (MHA) benefits more from increased expert numbers and ranks in bottom and top layers, whereas feed-forward networks (FFN) only exhibit this behavior in middle and top layers.

\section{Preliminary}
\paragraph{LoRA-MoE Framework}
LoRA-MoE integrates multiple vanilla LoRA experts into each pre-trained LLM submodule. Vanilla LoRA \cite{hu2022lora} efficiently adapts large models to downstream tasks by lowering computational and memory costs.
For a pre-trained weight matrix $\mathbf{W_0} \in \mathbb{R}^{m \times n}$, LoRA creates two low-rank trainable matrices $\mathbf{A}$ and $\mathbf{B}$, where $\mathbf{B} \in \mathbb{R}^{m \times r}$, $\mathbf{A} \in \mathbb{R}^{r \times n}$, where $r \ll min(m, n)$. During training, $\mathbf{W_0}$ remains fixed while $\mathbf{A}$ and $\mathbf{B}$ are updated via gradient descent. The output representation $h$ is defined as follows:
\begin{equation}
  \label{eq:lora}
  \mathbf{h} = \mathbf{W_0}x + \mathbf{BA}x
\end{equation}
\noindent Every traditional LoRA-MoE layer incorporates $N$ LoRA experts. The forward pass through the layer can be formulated as:
\begin{equation}
  \label{eq:mole}
  \mathbf{h} = \mathbf{W}_0x + \sum_{i=1}^{N} \mathbf{G}(x)_i\mathbf{B_i}\mathbf{A_i}x
\end{equation}
\noindent where $\mathbf{G}(x)=\text{Softmax}(x\mathbf{W_r})$ represents the router in the LoRA-MoE layer. $\mathbf{W}_r$ is the trainable parameter matrix of the routing network that directs input $x$ to different experts. By adaptively allocating inputs, the router promotes expert specialization, enhancing their ability to handle diverse tasks and input patterns.

\begin{figure*}[th]
  \includegraphics[width=2\columnwidth]{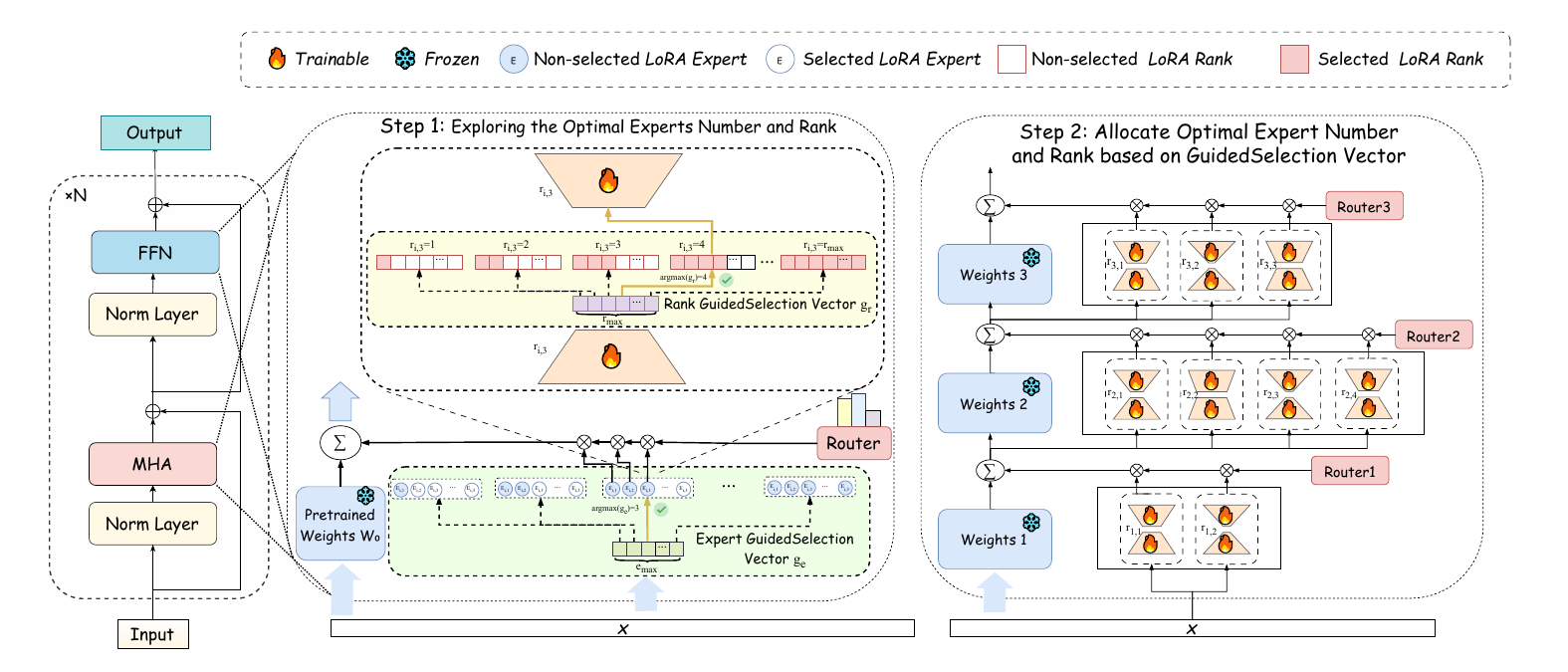}
  \caption{An illustration of our GuiLoMo strategy. GuiLoMo involves two steps: (Step 1): Exploring the optimal number of experts and ranks via a bilevel optimization algorithm with GuidedSelection Vectors. (Step 2): Allocate optimal expert number and rank based on GuidedSelection Vectors obtained in the previous step.}
  \label{fig:GuiLoMo}
\end{figure*}

\paragraph{Applying LoRA-MoE for LLMs} LoRA-MoE is applied to key modules of LLMs, namely multi-head attention (MHA) and feed-forward networks (FFNs). In MHA, inputs are projected via $\mathbf{W}^Q$, $\mathbf{W}^K$, $\mathbf{W}^V$, and $\mathbf{W}^O \in \mathbb{R}^{d \times d}$. Each FFN uses gate- and up-projection matrices $\mathbf{W}^G$, $\mathbf{W}^U \in \mathbb{R}^{d \times d'}$, a activation (e.g., GELU), and a down-projection $\mathbf{W}^D \in \mathbb{R}^{d' \times d}$, where $d' > d$. GuiLoMo assigns optimal expert number and rank to these matrices.

\section{Method}
In this section, we present our GuiLoMo strategy, which consists of two main steps: 1) A bilevel optimization algorithm is employed to obtain \textbf{GuidedSelection Vectors (GSVs)} of expert and rank for each module, tailored to the specific downstream task and model (\S~\ref{sec:bilevel_optimization}); 2) Based on the obtained GSVs, the optimal expert number and rank for each module in LoRA-MoE are allocated, and the final training is then conducted (\S~\ref{sec:Determining}). See \S~\ref{sec:expert_guidedselection} and \S~\ref{sec:rank_guidedselection} for details of GSVs. 

\subsection{Bilevel Optimization for Obtaining the GuidedSelection Vector}
\label{sec:bilevel_optimization}
In this section, we introduce the objective of bilevel optimization used to obtain \textbf{GuidedSelection Vectors} and its optimization process. 
\paragraph{Optimization Objective}
Formally, our objective is to automatically determine the optimal expert number $e^*_i$ for a given module (e.g., \texttt{down-projection in FFN}) within the $i$-th layer, and the optimal rank $r^*_{i,j}$ for the $j$-th expert under a specified LLM and downstream task setting. 

To achieve this, we formulate the problem as an optimization task. In this process, we introduce the Expert GuidedSelection Vector $\mathbf{g}_{\scriptscriptstyle E}$
and Rank GuidedSelection Vector $\mathbf{g}_{\scriptscriptstyle R}$ as key components of the optimization, and the optimization objective is:
\begin{align}
  \label{eq:optimization_objective}
  \min_{\{\mathbf{g}_{\scriptscriptstyle E}, \mathbf{g}_{\scriptscriptstyle R}\}} &\mathcal{L}(\mathcal{D}, \pi_{\theta}, \mathbf{g}_{\scriptscriptstyle E}, \mathbf{g}_{\scriptscriptstyle R}) \\
  \label{eq:LG}
  \mathcal{L} =  &\mathcal{L}_\text{SFT} + \mathcal{L}_\text{BAL}
\end{align}
\noindent where $\pi_\theta$ is specific LLM and $\mathcal{L}_\text{SFT}$ denotes the supervised fine-tuning loss, which is computed via autoregressive language modeling on the downstream dataset $\mathcal{D}$, while $\mathcal{L}_\text{BAL}$ (refer to Eq.~\ref{eq:BAL}) represents the MoE balancing loss \cite{fedus2022switch, zoph2022st}, which is introduced to encourage balanced utilization across experts and prevent expert collapse.
The GuidedSelection Vector $\mathbf{g}_{\scriptscriptstyle E} \in \mathbb{R}^{e_\text{max}}$ and $\mathbf{g}_{\scriptscriptstyle R} \in \mathbb{R}^{ r_\text{max}}$ are both trainable, with $e_\text{max}$ and $r_\text{max}$ representing the predefined maximum number of experts and ranks (see \S~\ref{sec:expert_guidedselection} and \S~\ref{sec:rank_guidedselection} for more details of $\mathbf{g}_{\scriptscriptstyle E}$ and $\mathbf{g}_{\scriptscriptstyle R}$). Since the optimization of $\mathbf{g}_{\scriptscriptstyle E}$, $\mathbf{g}_{\scriptscriptstyle R}$ should be under the optimal $\pi_{\theta}^*$, we draw inspiration from \citet{liu2018darts} and formulate the problem as a bilevel optimization:
\begin{equation}
  \label{eq:min_val_loss}
  \begin{aligned}
  &\min_{\{\mathbf{g}_{\scriptscriptstyle E}, \mathbf{g}_{\scriptscriptstyle R}\}} \mathcal{L}(\mathcal{D}_1, \pi_{\theta}^*, \mathbf{g}_{\scriptscriptstyle E}, \mathbf{g}_{\scriptscriptstyle R}) \\
  &\text{s.t.} \: \pi_{\theta}^* = \arg\min_{\pi_{\theta}} \: \mathcal{L}(\mathcal{D}_2, \pi_{\theta}, \mathbf{g}_{\scriptscriptstyle E}, \mathbf{g}_{\scriptscriptstyle R})
  \end{aligned} 
\end{equation}
\noindent where $\mathcal{D}_{1}$ and $\mathcal{D}_{2}$ are two splits of the training set $\mathcal{D}$ with equal size.
\paragraph{Optimization Process}
Based on the above objective, we formulate the overall procedure for obtaining the optimized GSVs in only a few $T$ training steps. For a specific $t$-th training step, we first obtain $\pi_{\theta}^*(t)$ following Eq.~\ref{eq:pi*}, and then optimize the GSVs $\mathbf{g}^{(t)}_{\scriptscriptstyle E / \scriptscriptstyle R} = \{\mathbf{g}_{\scriptscriptstyle E}, \mathbf{g}_{\scriptscriptstyle R}\}^{(t)}$ with $\pi_{\theta}^*(t)$ to obtain $\mathbf{g}^{(t+1)}_{\scriptscriptstyle E / \scriptscriptstyle R} = \{\mathbf{g}_{\scriptscriptstyle E}, \mathbf{g}_{\scriptscriptstyle R}\}^{(t+1)}$ following Eq.~\ref{eq:g*}.\footnote{$T$ is a hyperparameter in our experiments.} Finally, we use $\mathbf{g}^{(t+1)}_{\scriptscriptstyle E / \scriptscriptstyle R}$ to obtain $\pi_\theta(t+1)$ for next step following Eq.~\ref{eq:pi_next*}.

\vspace{-0.3cm}
{\small
\begin{flalign}
    \label{eq:pi*}
    \pi_{\theta}^*(t) =& \pi_{\theta}(t) - \xi_{\theta} * \nabla_{\pi_{\theta}(t)} \mathcal{L}(\mathcal{D}_{2}, \pi_{\theta}(t), \mathbf{g}^{(t)}_{\scriptscriptstyle E}, \mathbf{g}^{(t)}_{\scriptscriptstyle R}) \\
    \label{eq:g*}
    \mathbf{g}^{(t+1)}_{\scriptscriptstyle E / \scriptscriptstyle R} =& \mathbf{g}^{(t)}_{\scriptscriptstyle E / \scriptscriptstyle R} - \xi_{\mathbf{g}} * \hat{\nabla}_{\mathbf{g}^{(t)}_{\scriptscriptstyle E / \scriptscriptstyle R}} \mathcal{L}(\mathcal{D}_{1}, \pi_\theta^*(t), \mathbf{g}_{\scriptscriptstyle E}^{(t)}, \mathbf{g}_{\scriptscriptstyle R}^{(t)}) \\
    \label{eq:pi_next*}
    \pi_\theta(t+1) &= \pi_{\theta}(t) - \xi_{\theta} * \nabla_{\pi_{\theta}(t)} \mathcal{L}(\mathcal{D}_{2}, \pi_{\theta}(t), \mathbf{g}^{(t+1)}_{\scriptscriptstyle E}, \mathbf{g}^{(t+1)}_{\scriptscriptstyle R})
\end{flalign}
}

\noindent where $\xi_{\theta}$ and $\xi_{\mathbf{g}}$ are the learning rate for updating LLM weights and GSVs, respectively. $\hat{\nabla}$ indicates that we apply the STGE technique to ensure proper gradient flow (See \S~\ref{sec:expert_guidedselection} and \S~\ref{sec:rank_guidedselection} for more details).
The overall optimization procedure is summarized in Alg.~\ref{alg:algorithm1}. The final obtained $\mathbf{g}_{\scriptscriptstyle E}^*$ and $\mathbf{g}_{\scriptscriptstyle R}^*$ determine the optimal number of experts and ranks according to the strategy described in \S~\ref{sec:Determining}.
GuiLoMo progressively learns the optimal heterogeneous LoRA-MoE configuration, allowing it to meet model- and task-specific needs. 
\begin{algorithm}[th]
{
\caption{Optimization Process}
\label{alg:algorithm1}
\KwIn{Predefined maximum number of experts $e_\text{max}$ and LoRA rank $r_\text{max}$ per module, $T$ optimization steps, learning rates $\xi_{\theta}$ and $\xi_{\mathbf{g}}$ for Model weights and GSVs.}
\KwOut{The optimized Expert GSV $\mathbf{g}_{\scriptscriptstyle E}^*$ and Rank GSV $\mathbf{g}_{\scriptscriptstyle R}^*$.}  
    Initialize the LoRA-MoE framework according to the $e_\text{max}$ and $r_\text{max}$\;
    Split the training set $\mathcal{D}$ into $\mathcal{D}_{1}$ and $\mathcal{D}_{2}$\;
    \For{$t = 0$; $t < T$ }{
        Obtain LLM weight $\pi_\theta^*(t)$ using Eq.~\ref{eq:pi*} with learning rate $\xi_{\theta}$\;
        Obtain $\mathbf{g}_{\scriptscriptstyle E}^{(t+1)}$ and $\mathbf{g}_{\scriptscriptstyle R}^{(t+1)}$ using Eq.~\ref{eq:g*} with the gradients obtained from Eq.~\ref{eq:backward} and the learning rate $\xi_{\mathbf{g}}$\;
        Obtain LLM weight $\pi_\theta(t+1)$ using Eq.~\ref{eq:pi_next*} with the learning rate $\xi_{\theta}$\;
    }
    Derive the optimized Expert GSV $\mathbf{g}_{\scriptscriptstyle E}^*$ and Rank GSV $\mathbf{g}_{\scriptscriptstyle R}^*$.
}
\end{algorithm}

\subsection{Expert GuidedSelection Vector}
\label{sec:expert_guidedselection}
For the Expert GSVs $\mathbf{g}_{\scriptscriptstyle E} \in \mathbb{R}^{e_\text{max}}$, we first predefine the maximum expert number $e_\text{max}$ and initialize them with Gaussian distribution: 
\begin{equation}
  \label{eq:gE}
      \mathbf{g}_{\scriptscriptstyle E} = \text{Softmax}\!\bigl( \boldsymbol{\alpha} \bigr), \quad with\ \boldsymbol{\alpha} = \{\alpha_i\}_{i=1}^{e_\text{max}}
\end{equation}
\noindent where $\alpha_i \sim \mathcal{N}(0, 1)$, and $\mathbf{g}_{\scriptscriptstyle E}$ denotes the selection probabilities for different allocated expert number settings. GuiLoMo selects the expert number setting by taking the index of the maximum value in $\mathbf{g}_{\scriptscriptstyle E}$. For example, if the maximum value of $\mathbf{g}_{\scriptscriptstyle E}^i$ at the $i$-th layer occurs at the $3$-th position during the current training step, we allocate $3$ experts for this module (see the green region in Fig.~\ref{fig:GuiLoMo}).  
Since $\mathbf{g}_{\scriptscriptstyle E}^i$ is learned through a few optimization steps on the task-specific data, the expert selection process described above needs to be differentiable. To guarantee gradient flow and enable end-to-end  optimization, we adopt the Straight-Through Gradient Estimator (STGE) \cite{bengio2013estimating} along with an auxiliary virtual vector $\mathcal{M}_E$ to approximate discrete selection while maintaining differentiability. 
Let $n^{\star}$ denote the index of the maximum value in $\mathbf{g}_{\scriptscriptstyle E}$. The forward propagation of the expert virtual vector $\mathcal{M}_E\in \{0,-\infty\}^{e_\text{max}}$ is formulated as follows:
\begin{equation}
    \label{eq:forward_E}
    \mathcal{M}_E^i =
    \left\{
    \begin{array}{ll}
    0, & \text{if } i \leq n^{\star} \\
    -\infty, & \text{if } i > n^{\star}
    \end{array}
    \right.
\end{equation}
For example, when allocating $3$ experts, the expert virtual vector $\mathcal{M}_E$ is: $[0,0,0,-\infty,...,-\infty]$. 
Meanwhile, in the backward propagation,
we propagate the gradient flow from $\mathcal{M}_E$ to $\mathbf{g}_{\scriptscriptstyle E}$:
\begin{equation}
    \label{eq:backward}
    \frac{\partial \mathcal{L}}{\partial \mathbf{g}_{\scriptscriptstyle E}} = \mathcal{H}(\frac{\partial \mathcal{L}}{\partial \mathcal{M}_E} )
\end{equation}
\noindent For more details on the $\mathcal{H}$ operation, please refer to Appendix~\ref{section:gradientg}.
The $\mathcal{M}_E$ is applied to top-K routing process to guide the learning of $\mathbf{g}_{\scriptscriptstyle E}$:

\vspace{-0.3cm}
{\small
\begin{equation}
      \hat{G}(x) = \frac{\text{TopK}(\text{Softmax} (x\mathbf{W}_{r}+\mathcal{M}_E), K)_i} {\sum_{i=1}^{K}\text{TopK}(\text{Softmax} (x\mathbf{W}_{r}+\mathcal{M}_E), K)_i} 
  \label{eq:Gx}
\end{equation}
}
\noindent where $\mathbf{W}_{r}$ denotes the weight of routing network. 
\subsection{Rank GuidedSelection Vector}
\label{sec:rank_guidedselection}
The Rank GSVs $\mathbf{g}_{\scriptscriptstyle R\in\mathbb{R}^{r_{\max}}}$ shares a similar concept with the Expert GSVs during bilevel optimization. It begins by predefining the maximum rank $r_{\text{max}}$ and is also initialized with Gaussian distribution using Eq.~\ref{eq:gE}. However, the semantic meaning of each element differs, where each element in $\mathbf{g}_{\scriptscriptstyle R}$ represents a specific rank assigned to the corresponding expert.
We select the index of maximum value in $\mathbf{g}_{\scriptscriptstyle R}$, i.e., $m^{\star}$, to determine the rank for the current training step. 
Similar to $\mathbf{g}_{\scriptscriptstyle E}$, $\mathbf{g}_{\scriptscriptstyle R}$ is non-differentiable during this process; therefore, we design rank virtual vector $\mathcal{M}_R \in \{0,1\}^{r_\text{max}}$ to address this issue: 
\begin{equation}
    \label{eq:forward_R}
    \mathcal{M}_R^i =
    \left\{
    \begin{array}{ll}
    1, & \text{if } i \leq m^{\star} \\
    0, & \text{if } i > m^{\star}
    \end{array}
    \right.
\end{equation}

\begin{table*}[th]
  \centering
  \setlength\tabcolsep{2pt}
  \fontsize{9.5}{10}\selectfont 
    \begin{tabular}{l|l|ccc|ccc|c}
          \toprule
          \toprule
          Models & Strategy & MRPC & COLA & RTE & ScienceQA & CommonsenseQA & OpenBookQA & Avg. \\
          \midrule
          \multirow{5}{*}{$\text{LLaMA}_\text{7B}$} 
          & MoLA(5)-Uniform(8) & 82.43 & 84.18 & 83.03 & 90.28 & 75.10 & 76.00 & 81.84 \\
          & AlphaLoRA-Uniform(8) & {85.19} & 85.42 & 85.19 & 90.37 & 76.49 & 78.20 & 83.48 \\
          & MoLA(5)\ +\ SoRA & 82.55 & 84.76 & 83.03 & 90.38 & 75.35 & 76.80 & 82.15 \\
          & AlphaLoRA\ +\ SoRA & \textbf{85.51} & 85.62 & 85.20 & 90.78 & 76.82 & 78.20 & 83.69 \\
          & GuiLoMo (Ours) & 85.04 & \textbf{85.71} & \textbf{85.92} & \textbf{91.50} & \textbf{77.15} & \textbf{78.60} & \textbf{83.99} \\
          \midrule
  
          \multirow{5}{*}{$\text{LLaMA-2}_\text{7B}$} 
          & MoLA(5)-Uniform(8) & 84.17 & 86.19 & 84.83 & 92.08 & 77.55 & 80.00 & 84.14  \\
          & AlphaLoRA-Uniform(8) & 84.23 & 86.67 & \textbf{87.36} & 92.71 & 78.05 & 80.80 & 84.97 \\
          & MoLA(5)\ +\ SoRA & 84.46 & 86.31 & 84.84 & 92.36 & 77.81 & 80.20 & 84.31 \\
          & AlphaLoRA\ +\ SoRA & 84.99 & 85.81 & 87.00 & 92.31 & 78.38 & 80.00 & 84.75 \\
          & GuiLoMo (Ours) & \textbf{85.80} & \textbf{87.25} & \textbf{87.36} & \textbf{92.99} & \textbf{78.46} & \textbf{81.20} & \textbf{85.51} \\
          \midrule
          
          \multirow{5}{*}{$\text{LLaMA-3}_\text{8B}$} 
          & MoLA(5)--Uniform(8) & 86.61 & 87.15 & 87.73 & 93.97 & 79.52 & 83.40 & 86.40 \\
          & AlphaLoRA-Uniform(8) & 87.13 & 88.88 & 88.09 & 94.42 & 80.02 & 83.80 & 87.06 \\
          & MoLA(5)\ +\ SoRA & 85.97 & 87.54 & 88.45 & 94.24 & 79.44 & 84.00 & 86.61 \\
          & AlphaLoRA\ +\ SoRA & 87.07 & 88.69 & \textbf{89.53} & 94.20 & 80.18 & 84.00 & 87.28 \\
          & GuiLoMo (Ours) & \textbf{87.77} & \textbf{89.26} & {88.45} & \textbf{94.83} & \textbf{81.24} & \textbf{85.60} & \textbf{87.86} \\
          \bottomrule
          \bottomrule
      \end{tabular}
    \caption{\label{table1}
      Accuracy comparison of different methods under direct fine-tuning for each dataset. MoLA(5) indicates assigning a uniform 5 experts to each layer. Uniform(8) denotes setting all the rank of LoRA expert to 8.
    }
  \end{table*}

\noindent For example, if the maximum value of $\mathbf{g}_{\scriptscriptstyle R}$ at a given training step is located at the $4$-th element, the rank for this module is set to $4$ (see the yellow region in Fig.~\ref{fig:GuiLoMo}). Accordingly, the corresponding Rank GuidedSelection Vector $\mathcal{M}_R$ is $[1,1,1,1,0,...,0]$.

Then, we parameterize on each LoRA expert matrix, denoted as $\Delta=\mathbf{BA} \in \mathbb{R}^{m \times n}$ (Eq.~\ref{eq:lora}), in a form that mimic singular value decomposition (SVD) to obtain $\Delta = \mathbf{P \Lambda Q}$.
$\mathbf{P} \in \mathbb{R}^{d_1 \times r_\text{max}}$ and $\mathbf{Q}\in\mathbb{R}^{r_\text{max} \times d_2}$ correspond to the original LoRA matrices $\mathbf{B}$ and $\mathbf{A}$, respectively, and $\mathbf{\Lambda}$ are initialized to $1$. Note that we do not perform exact SVD. Subsequently, the rank virtual vector $\mathcal{M}_R$ is integrated with $\mathbf{\Lambda}$ and is incorporated into Eq.~\ref{eq:mole} to perform forward propagation:
\begin{equation}
  \label{eq:h}
  \mathbf{h}' = \mathbf{W}_0x + \sum_{i=1}^{K} \mathbf{\hat{G}}(x)_i\mathbf{P}  (\mathcal{M}_R \odot \mathbf{\Lambda} \odot \mathbf{Q}x ) 
\end{equation}
\noindent where $\odot$ denotes element-wise dot product, and $\mathbf{\hat{G}}$ is defined in Eq.~\ref{eq:Gx}.
$\mathcal{M}_R$ guide the learning of $\mathbf{g}_{\scriptscriptstyle R}$, and its gradients are backpropagated in the same manner as $\mathcal{M}_E$ in Eq.~\ref{eq:backward}, using STGE technique. 

\subsection{Allocating Expert Number and Rank via GSV}
\label{sec:Determining}
After obtaining optimized expert and rank GSVs, i.e., $\mathbf{g}_{\scriptscriptstyle E}^*$ and $\mathbf{g}_{\scriptscriptstyle R}^*$, the optimal expert number $e^*$ and rank $r^*$ are determined by selecting the index corresponding to the maximum values:
\begin{equation}
  \label{eq:argmax}
  \begin{aligned}
      e^*_i &= \text{argmax}\ (\mathbf{g}_{\scriptscriptstyle E}^*i) \\
      r^*_{i,j} &= \text{argmax}\ (\mathbf{g}_{\scriptscriptstyle R}^{*i,j}) \\
  \end{aligned} 
\end{equation}
\noindent where $e^*_i \leq e_\text{max}$ and $r^*_{i,j} \leq r_\text{max}$ denote the assigned expert number and rank in the $i$-th layer and the rank of the $j$-th expert in the $i$-th layer, respectively.
Subsequently, we fine-tune the model using the loss function defined in Eq.~\ref{eq:LG} with expert number $e^*$ and rank $r^*$, where the LoRA-MoE weights are initialized with $\pi_{\theta}^{*}(T)$.

\section{Experiment}
In this section, we conduct extensive experiments to examine the performance of GuiLoMo. 
We also conduct extra experimental analyses to gain deeper insights into this field, as presented in \S~\ref{sec:Further analysis}. Implementation details can be
found in Appendix~\ref{sec:Implementation}.

\subsection{Experimental Settings}
\paragraph{Datasets}
Following \citet{qing-etal-2024-alphalora}, we evaluate our model on three natural-language understanding (NLU) tasks from GLUE: (1) the Microsoft Research Paraphrase Corpus (MRPC) \cite{dolan2005automatically},
(2) the Recognizing Textual Entailment (RTE) dataset \cite{wang2018glue},
(3) the Corpus of Linguistic Acceptability (CoLA) \cite{wang2018glue}, and three reasoning-focused question-answering (QA) tasks:
(1) ScienceQA \cite{lu2022learn},
(2) CommonsenseQA \cite{talmor-etal-2019-commonsenseqa}, and
(3) OpenBookQA \cite{mihaylov-etal-2018-suit}.

We also evaluate GuiLoMo on mathematical reasoning benchmarks. Specifically, we perform instruction tuning on the MetaMathQA \cite{yu2024metamath} dataset and evaluate on three benchmarks: (1) MultiArith \cite{roy2015reasoning}, (2) SVAMP \cite{patel-etal-2021-nlp}, and (3) GSM8K \cite{cobbe2021training}.
See Appendix \ref{section:Dataset} for the detailed statistics of all the datasets used in our experiments.
\paragraph{Models}
We have applied our method to $\text{LLaMA}_\text{7B}$ \cite{touvron2023llamaopenefficientfoundation}, $\text{LLaMA-2}_\text{7B}$ \cite{touvron2023llama}, $\text{LLaMA-3}_\text{8B}$ \cite{cobbe2021training}, and $\text{Mistral-v0.1}_\text{7B}$ \cite{jiang2023mistral7b}.  

\paragraph{Baselines}
We compare our GuiLoMo strategy with current state-of-the-art (SOTA) methods including MoLA \cite{gao2024higher}, AlphaLoRA \cite{qing-etal-2024-alphalora}, MoLA+SoRA, and AlphaLoRA+SoRA. SoRA \cite{ding-etal-2023-sparse} is a variant of LoRA that allows for dynamic adjustments to the intrinsic rank during the adaptation process.\footnote{We adopt SoRA as it represents the current SOTA among LoRA-based methods that enable dynamic adjustments to the intrinsic rank during the adaptation process.}
Implementation details of baselines can be found in Appendix~\ref{section:baselines}.

\subsection{Main Result}
\begin{table}[!t]
  \centering
  \setlength\tabcolsep{2.5pt}
\fontsize{7.5}{9}\selectfont 
  \begin{tabular}{l|l|ccc|c}
        \toprule
        \toprule
        Models & Strategy & GSM8K & SVAMP & MultiArith & Avg. \\
        \midrule
        \multirow{5}{*}{$\text{LLaMA}_\text{7B}$} 
        & M(5)-U(8) & 44.04 & 52.00 & 88.16 & 61.40 \\
        & A-U(8) & 45.03 & 53.60 & 86.67 & 61.77 \\
        & M(5)\ +\ S & 43.97 & 53.80 & 88.50 & 62.09 \\
        & A\ +\ S & 46.10 & 54.80 & 89.17 & 63.36 \\
        
        & GuiLoMo (Ours) & \textbf{47.01} & \textbf{56.60} & \textbf{91.17} & \textbf{64.93} \\
        \midrule
        \multirow{5}{*}{$\text{LLaMA-2}_\text{7B}$} 
        & M(5)-U(8) & 49.50 & 57.10 & 87.00 & 64.53  \\
        & A-U(8) & 50.42 & 57.00 & 91.33 & 66.25 \\
        & M(5)\ +\ S & 50.42 & 57.90 & 88.33 & 65.55 \\
        & A\ +\ S & 51.48 & 58.00 & 92.50 & 67.33 \\
        
        & GuiLoMo (Ours) & \textbf{53.07} & \textbf{59.20} & \textbf{93.67} & \textbf{68.65} \\
        \midrule
        \multirow{5}{*}{$\text{LLaMA-3}_\text{8B}$} 
        & M(5)-U(8) & 71.03 & 74.30 & 96.83 & 80.72  \\
        & A-U(8) & 71.49 & 75.10 & 97.33 & 81.31 \\
        & M(5)\ +\ S & 71.72 & 73.50 & 97.00 & 80.74 \\
        & A\ +\ S & \textbf{73.01} & 75.30 & 97.33 & 81.88 \\
        
        & GuiLoMo (Ours) & 72.85 & \textbf{76.00} & \textbf{97.83} & \textbf{82.23} \\
        \bottomrule
        \bottomrule
    \end{tabular}
  \caption{\label{table2}
    The results of mathematical reasoning under three models. 
    M(5)-U(8) denotes MoLA(5)--Uniform(8); A-U(8) denotes AlphaLoRA-Uniform(8); M(5) + S denotes MoLA(5) + SoRA; A +S: AlphaLoRA + SoRA. MoLA(5) indicates assigning a uniform 5 experts to each layer. Uniform(8) represents setting all the rank of LoRA expert to 8.
  }
  \vspace{-0.2cm}
\end{table}

Table \ref{table1} reports the results on three NLU tasks and three QA benchmarks.
Across these datasets, GuiLoMo surpasses every baseline in terms of Avg. performance. 
Specifically, relative to AlphaLoRA-Uniform(8)\footnote{``Uniform(8)'' refers to assigning the same rank 8 to all LoRA experts.}, GuiLoMo delivers consistent gains of $0.61\%$, $0.64\%$, and $0.84\%$ on the three model settings, respectively.
GuiLoMo also outperform baselines on mathematical reasoning task.
As show in Table~\ref{table2}, GuiLoMo 
exceeds AlphaLoRA + SoRA by an average of of $2.48\%$, $2.61\%$, and $0.43\%$ on $\text{LLaMA}_\text{7B}$, $\text{LLaMA-2}_\text{7B}$, and $\text{LLaMA-3}_\text{8B}$, respectively. 
Based on these observations, we conclude that: \textit{GuiLoMo, which flexibly allocates expert number and rank tailored to both model- and task-specific demands, further unleashes the potential of LoRA-MoE and leads to improved performance.}

\subsection{Further Analysis}
\label{sec:Further analysis}
\paragraph{Ablation Study of GuiLoMo Strategy}
\begin{table}[!t]
  \centering
  \setlength\tabcolsep{5pt}
\fontsize{10}{10}\selectfont 
      \begin{tabular}{l|cccccc|c}
            \toprule
            \toprule
            Strategy & Avg. \\
            \midrule
            MoLA(5)-Uniform(8) & 79.42 \\
            \midrule 
            GuiLoMo (Ours) & \textbf{81.87} \\
            w/o adaptive expert allocation & 80.64 \\
            w/o varying rank & 80.97 \\
            \bottomrule
            \bottomrule
        \end{tabular}
      \caption{\label{ablation}
        Average results of ablation studies on GuiLoMo across six tasks. MoLA(5) indicates assigning a uniform 5 experts to each layer. Uniform(8) represents setting all the rank of LoRA expert to 8. See Table~\ref{detailed_ablation} for detailed results. ``w/o''  means the exclusion of this strategy from GuiLoMo. 
      }
\end{table}

We conduct ablation studies to assess the effectiveness of GuiLoMo with $\text{LLaMA-2}_\text{7B}$ across NLU and QA benchmarks on two different settings:
(1) a fixed uniformly distributed number of experts with varying ranks, (2) a fixed uniformly assigned rank with varying expert allocation. As shown in Table~\ref{ablation}, compared with the uniformly-allocated baseline MoLA(5)-Uniform(8), applying GuiLoMo exclusively to expert allocation or exclusively to rank allocation results in average performance improvements of $1.95\%$ and $1.53\%$, respectively. 
The results also show that excluding either expert allocation or rank allocation from GuiLoMo leads to performance drops of $1.50\%$ and $1.10\%$, respectively. Accordingly, we highlight the following insight:
\begin{tcolorbox}
\vspace{-0.2cm}
\textbf{Insight 1.} Jointly optimizing both expert and rank allocations outperforms optimizing either one in isolation.
\vspace{-0.2cm}
\end{tcolorbox}

\paragraph{Results across Model Families and Scales}
We conduct extra experiments on another family model $\text{Mistral-v0.1}_\text{7B}$ and larger-scale model $\text{LLaMA-2}_\text{13B}$ across three benchmarks to examine the generalization of our GuiLoMo.
As shown in Table~\ref{families_scales}, GuiLoMo achieves average score improvements of $0.79\%$ and $0.18\%$ over the AlphaLoRA+SoRA on $\text{LLaMA-2}_\text{13B}$ and $\text{Mistral-v0.1}_\text{7B}$, respectively.
The results further validate the widespread effectiveness of GuiLoMo across models of different scales and families.

\begin{table}[!t]
  \centering
  \setlength\tabcolsep{1pt}
\fontsize{8}{9}\selectfont 
      \begin{tabular}{llccccc}
            \toprule
            \toprule
            Models & Strategy & MRPC & COLA & ComQA & Avg. \\
            \midrule
            \multirow{3}{*}{$\text{LLaMA-2}_\text{13B}$} 
            & MoLA(5)–Uniform(8) & 86.78 & 87.82 & 81.74 & 85.45 \\
            & AlphaLoRA + SoRA & 87.13 & 88.97 & 83.21 & 86.44 \\
            & GuiLoMo (Ours)  & \textbf{88.06} & \textbf{89.36} & \textbf{83.95} & \textbf{87.12} \\
            \midrule
    
            \multirow{3}{*}{$\text{Mistral-v0.1}_\text{7B}$} 
            & MoLA(5)–Uniform(8) & 86.43 & 87.24 & 82.96 & 85.54 \\
            & AlphaLoRA + SoRA & 88.00 & \textbf{89.26} & 83.87 & 87.04 \\
            & GuiLoMo (Ours)  & \textbf{88.23} & 89.17 & \textbf{84.19} & \textbf{87.20} \\
            \bottomrule
            \bottomrule
        \end{tabular}
\vspace{-0.1cm}
  \caption{\label{families_scales}
    The scores on MRPC, COLA, and ComQA under the $\text{Mistral-v0.1}_\text{7B}$ and $\text{LLaMA-2}_\text{13B}$ models. Avg.: the average score over these three benchmarks.
  }
\end{table}

\paragraph{Effectiveness of the Expert Number and Rank Assigned by GuiLoMo}
To validate the effectiveness of expert number $e^*$ and rank $r^*$ assigned by GuiLoMo that is tailored to specific models and tasks, we additionally conduct experiments with the following three strategies using $\text{LLaMA-2}_\text{7B}$ on COLA benchmark.
\textbf{1) \underline{I}ncrease in \underline{E}xpert \underline{N}umber (IEN) }, increasing the number of experts while keeping the total rank ($\sum_{i=1}^{N}\sum_{j=1}^{e_{i}^*}{r^*_{i,j}}$) constant; 
\textbf{2) \underline{D}ecrease in \underline{E}xpert \underline{N}umber (DEN)}: Decreasing the number of experts while keeping the total rank constant~\footnote{To maintain a constant total rank in the LoRA-MoE framework, we proportionally reduce (or increase) the rank $r^*$ previously assigned by GuiLoMo to each individual LoRA expert when total number of experts is increased (or decreased).};
\textbf{3) \underline{M}ixed \underline{R}ank \underline{A}djustment (MRA) }: Keeping the number of experts fixed, we randomly reassign ranks while keeping the total rank unchanged.

\begin{figure}[t]
  \includegraphics[width=\columnwidth]{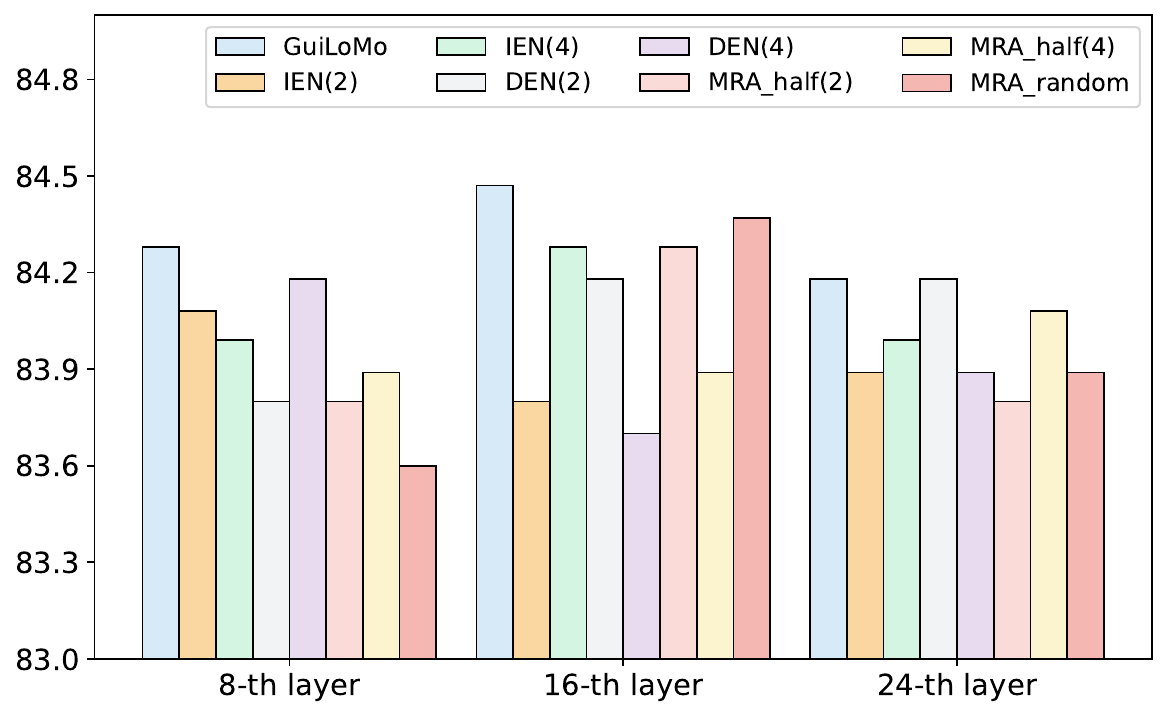}
  \vspace{-0.6cm}
  \caption{A comparative study of perturbed expert number $e^*$ and rank $r^*$ at different layers ($8$-th, $16$-th, and $24$-th). IEN(*) and DEN(*) denote the addition and removal of * experts, respectively. MRA\_half(*): Half of the LoRA experts have their ranks increased by *, while the other half have their ranks decreased by * accordingly. MRA\_random: Randomly shuffling the ranks of LoRA experts.}
  \label{fig:grouped_bar_chart_pastel}
  \vspace{-0.3cm}
\end{figure}

Note that only the expert number and rank of the specific $m$-th layer are intervened using the above three strategies, while the expert number and rank of the remaining layers remain unchanged (allocated by GuiLoMo). We apply these strategies to three layers (8, 16, 24) and report the results in Fig.~\ref{fig:grouped_bar_chart_pastel}.
The results show that GuiLoMo outperforms all modified configurations, achieving the highest overall performance. From the results, we distill the following insight:
\begin{tcolorbox}
\vspace{-0.2cm}
\textbf{Insight 2.} GuiLoMo allocates layer-wise optimal expert number and rank, better exploiting the potential of LoRA-MoE.
\vspace{-0.2cm}
\end{tcolorbox}

\paragraph{Allocation for MHA and FFN}
\begin{figure}[t]
  \includegraphics[width=\columnwidth]{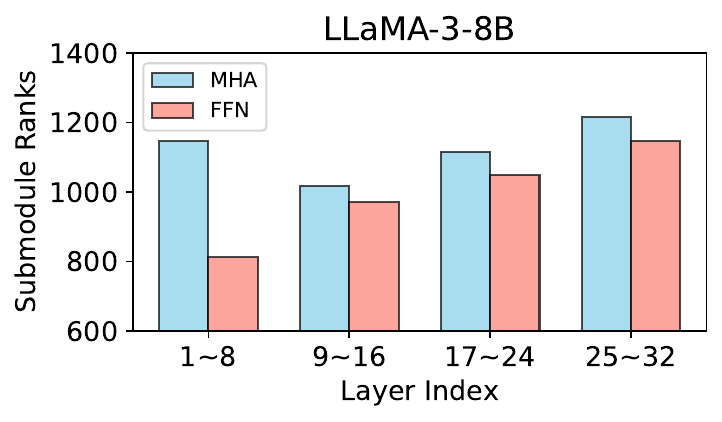}
  \vspace{-0.6cm}
  \caption{Total Rank of sub-modules (MHA and FFN) across different layer ranges in $\text{LLaMA-3}_\text{8B}$ on CommonsenseQA.}
  \vspace{-0.3cm}
  \label{fig:bar_chart-Llama-3-8B-grouped}
\end{figure}
\begin{figure}[t]
  \includegraphics[width=\columnwidth]{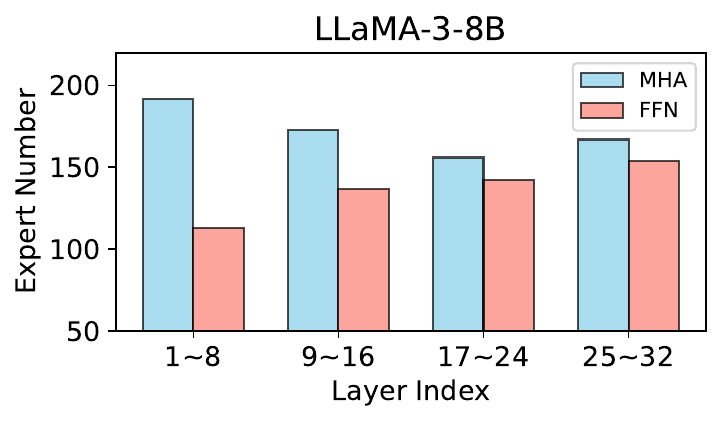}
  \vspace{-0.6cm}
  \caption{Total number of allocated experts for sub-modules (MHA and FFN) across different layer ranges in $\text{LLaMA-3}_\text{8B}$ on CommonsenseQA.}
  \vspace{-0.3cm}
  \label{fig:bar_chart-Llama-3-8B-expert}
\end{figure}
To delve deeper, we also observe the allocation patterns for MHA and FFN separately. 
We report the total assigned rank and average number of allocated experts for MHA and FFN under different layer ranges in Fig.~\ref{fig:bar_chart-Llama-3-8B-grouped} and Fig.~\ref{fig:bar_chart-Llama-3-8B-expert}, respectively.
For example, the total rank ($\text{Total Rank of Submodules} = (\sum_{i=1}^8\sum_{j=1}^{e^*_i}r^*_{i,j})$) in layer range $1\sim8$ of FFN, which includes gate-, up-, and down-projection.
Based on Fig.~\ref{fig:bar_chart-Llama-3-8B-grouped} and Fig.~\ref{fig:bar_chart-Llama-3-8B-expert}, we draw the conclusion (see similar observations on other models and tasks in Appendix~\ref{section:Parameter_other}): 
\begin{tcolorbox}
\vspace{-0.2cm}
\textbf{Insight 3.} MHA requires more expert numbers and ranks in bottom and top layers, whereas FFN shows this trend mainly in the middle and top layers.

\vspace{-0.2cm}
\end{tcolorbox}
\paragraph{Expert diversity}
We also explore \underline{E}xpert \underline{D}iversity (ED) by quantifying it as the ratio between the size of the largest subset of experts whose ranks are all mutually distinct and the total number of experts ($\text{ED} = |\text{largest\ rank-distinct\ subset}|\ /\ |\text{all\ experts}|$).

For example, consider the FFN's up-projection module, which contains five experts with ranks 
$[3,5,6,3,7]$, so the expert diversity score $\text{ED}=4/5=0.8$. 
In Fig.~\ref{fig:pie_chart}, 
we analyze the ED score for each submodule across NLU benchmarks on $\text{LLaMA}_\text{7B}$, $\text{LLaMA-2}_\text{7B}$, and $\text{LLaMA-3}_\text{8B}$. The results show that $38.1\%$ of the ED scores fall within the high range of $0.75\sim1.00$, whereas only 8.7\% are in the low range of $0.00\sim0.25$. Based on this observation, we draw the following conclusion: 
\begin{figure}[t]
  \includegraphics[width=0.95\columnwidth]{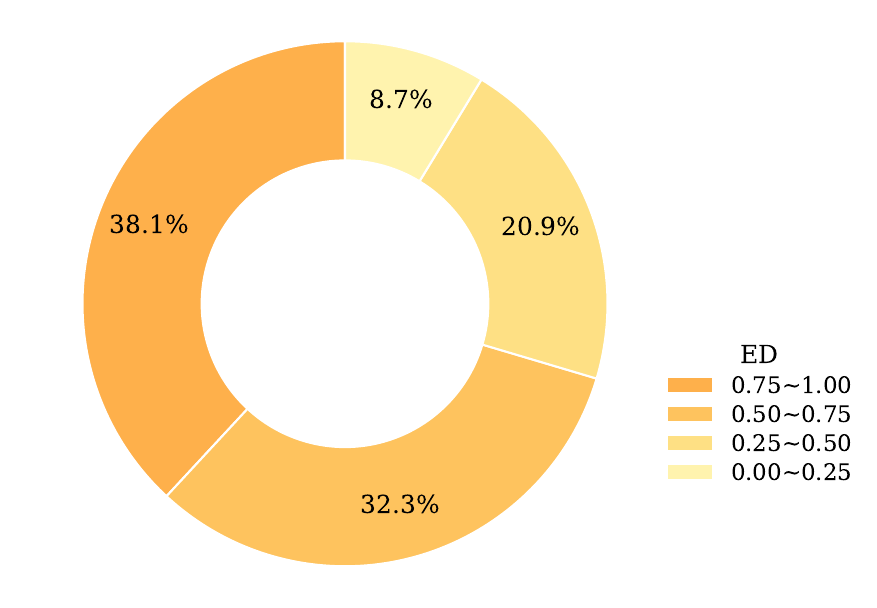}
    \vspace{-0.3cm}
    \caption{Distribution of ED scores computed by all the modules on $\text{LLaMA}_\text{7B}$, $\text{LLaMA-2}_\text{7B}$, and $\text{LLaMA-3}_\text{8B}$ under three NLU tasks.}
    \label{fig:pie_chart}
  \end{figure}

\begin{tcolorbox}
\vspace{-0.2cm}
\textbf{Insight 4.} Allocating diverse expert ranks enables more flexible and specialized adaptation to different tasks.
\vspace{-0.2cm}
\end{tcolorbox}

\paragraph{Impact of Task Difficulty}
We aim to investigate how the expert number $e^*$ and rank $r^*$ derived by GuiLoMo differ when facing challenging tasks compared to simpler ones.
In pursuit of this goal, we use two BBH \cite{suzgun-etal-2023-challenging} sub-tasks, \textbf{Tracking Shuffled Objects} and \textbf{Logical Deduction}\footnote{Given a set of $K$ objects with initial positions and a sequence of pairwise swaps, determine their final positions after all transformations. Determine the sequential arrangement of $K$ objects based on provided information regarding their relative positions and spatial relationships.}, each consists of sub-tasks differing in the number of objects $K$ involved, with difficulty increasing as the number of objects grows.

From Table~\ref{table5}, we observe that as the number of objects increases, the number of experts assigned to different sub-tasks scales proportionally with the number of elements. However, the rank does not exhibit such a trend. Hence, we derive the following insight: 
\begin{tcolorbox}
\vspace{-0.2cm}
\textbf{Insight 5.} Within the LoRA-MoE, harder tasks benefit more from adding experts than from raising the rank of each LoRA expert.
\vspace{-0.5cm}
\end{tcolorbox}

\begin{table}[t]
    \centering
    \setlength\tabcolsep{2.5pt}
\fontsize{9}{9}\selectfont 
        \begin{tabular}{lcc}
            \toprule
            \toprule
            Task & Avg. Expert & Avg. Rank  \\
            \midrule
            Tracking shuffled objects & \\
            \midrule
            \specialrule{0em}{1pt}{1pt}
            --- Object Number $\text{K} = \text{3}$ & 5.87 & 6.13 \\
            --- Object Number $\text{K} = \text{5}$ & 6.13 & 5.98 \\
            --- Object Number $\text{K} = \text{7}$ & 6.35 & 6.07 \\
            \midrule
            Logical deduction & \\
            \midrule
            \specialrule{0em}{1pt}{1pt}
            --- Object Number $\text{K} = \text{3}$ & 5.23 & 6.44 \\
            --- Object Number $\text{K} = \text{5}$ & 5.41 & 6.51 \\
            --- Object Number $\text{K} = \text{7}$ & 5.76 & 6.40 \\
            \bottomrule
            \bottomrule
        \end{tabular}
    \caption{\label{table5}
   The average number of assigned experts across all modules (``module'' here refers to all weight matrices where LoRA-MoE is applied, i.e., $\mathbf{W}^Q$, $\mathbf{W}^K$, $\mathbf{W}^V$, $\mathbf{W}^O$ in MHA and $\mathbf{W}^U$, $\mathbf{W}^D$, $\mathbf{W}^G$ in FFN) and the average rank across all experts, calculated under different object numbers within the same task.}
\end{table}

\section{Related work}
\paragraph{LoRA-MoE Framework}
Recent research explores the integration of
MoE \cite{shazeer2017outrageously} and LoRA~\citep{hu2022lora}, referred to as LoRA-MoE, to
boost performance of LLMs in both single-task and multi-task scenarios in an efficient manner~\cite{wu2024mixture,gao2024higher,qing-etal-2024-alphalora,dou-etal-2024-loramoe,liu2023moelora,luo2024moelora}. For instance, \citet{dou-etal-2024-loramoe} leveraged LoRA-MoE to reduce catastrophic forgetting problem during supervised fine-tuning. \citet{wu2024mixture} uitlized LoRA-MoE with a hierarchical gating mechanism for efficient fusion across NLP and Vision \& Language tasks. However, existing works merely allocate expert numbers and ranks uniformly for LoRA-MoE, failing to fully exploit its potential.

\paragraph{Allocation Strategy for LoRA-MoE}
To exploit the potential of LoRA-MoE, \citet{gao2024higher} revealed that higher layers require more LoRA experts and initialized LoRA-MoE with different numbers of experts with group-wise allocations. Moreover, \citet{qing-etal-2024-alphalora} leveraged Heavy-Tailed Self-Regularization (HT-SR) theory to develop a training-free and theoretically grounded method for allocating suitable expert numbers for LoRA-MoE. However, previous methods only consider the expert number while overlooking the expert rank, which results in all experts having the same capacity and thus lacking diversity. In contrast, our proposed GuiLoMo jointly optimizes both the expert number and rank.

\section{Conclusion}
\label{sec:conclusion}
In this work, we propose GuiLoMo, a fine-grained allocation strategy designed to fully exploit the potential of LoRA-MoE. GuiLoMo jointly determines expert number and rank through a bilevel optimization process. Unlike prior methods that rely on uniform or task-agnostic configurations, it introduces a GuidedSelection mechanism that guides the layer-wise allocation of expert number and rank in LoRA-MoE, tailored to both model-specific and task-specific needs. Extensive experiments demonstrate that GuiLoMo consistently improves model performance across a wide range of tasks. Furthermore, our analysis reveals how optimal expert configurations vary across layers and tasks, offering deeper insights into this field. We believe GuiLoMo paves the way for more flexible and efficient expert allocation strategies in future research.

\section*{Acknowledgement}
This work is supported by the National Key R\&D Program of China (2023YFC3304902). We would like to thank all the anonymous reviewers for their valuable comments and constructive feedback.

\section*{Limitations}
While GuiLoMo demonstrates strong effectiveness and scalability in allocating expert number and rank for LoRA-MoE to both model-specific and task-specific settings, there remain two limitations. First, our experiments are limited to models up to 13B parameters, and we have not evaluated GuiLoMo on larger open-source LLMs such as LLaMA-70B due to computational constraints. Exploring its behavior on such super-sized models may provide further insights into scalability. Second, our evaluation is restricted to standard NLP tasks. It remains unclear whether GuiLoMo generalizes to other modalities or task types, such as cross-modal or multi-modal scenarios. We leave these directions for future work.

\bibliography{references}

\newpage
\appendix

\newpage\onecolumn
\section{Effect of Expert Number and Rank on Diverse Downstream Tasks
}
\label{sec:Diverse_Downstream}
\begin{figure*}[h]
\includegraphics[width=\textwidth]{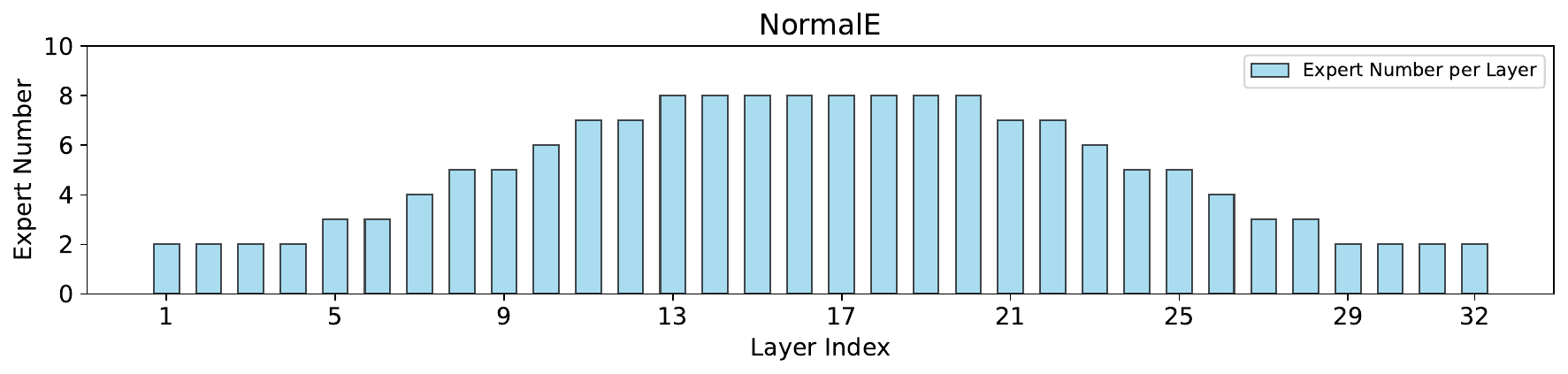}
  \caption{Allocating expert number across different layers using NormalE strategy.}
  \label{fig:NormalE}
\end{figure*}

\begin{figure*}[h]
  \includegraphics[width=\textwidth]{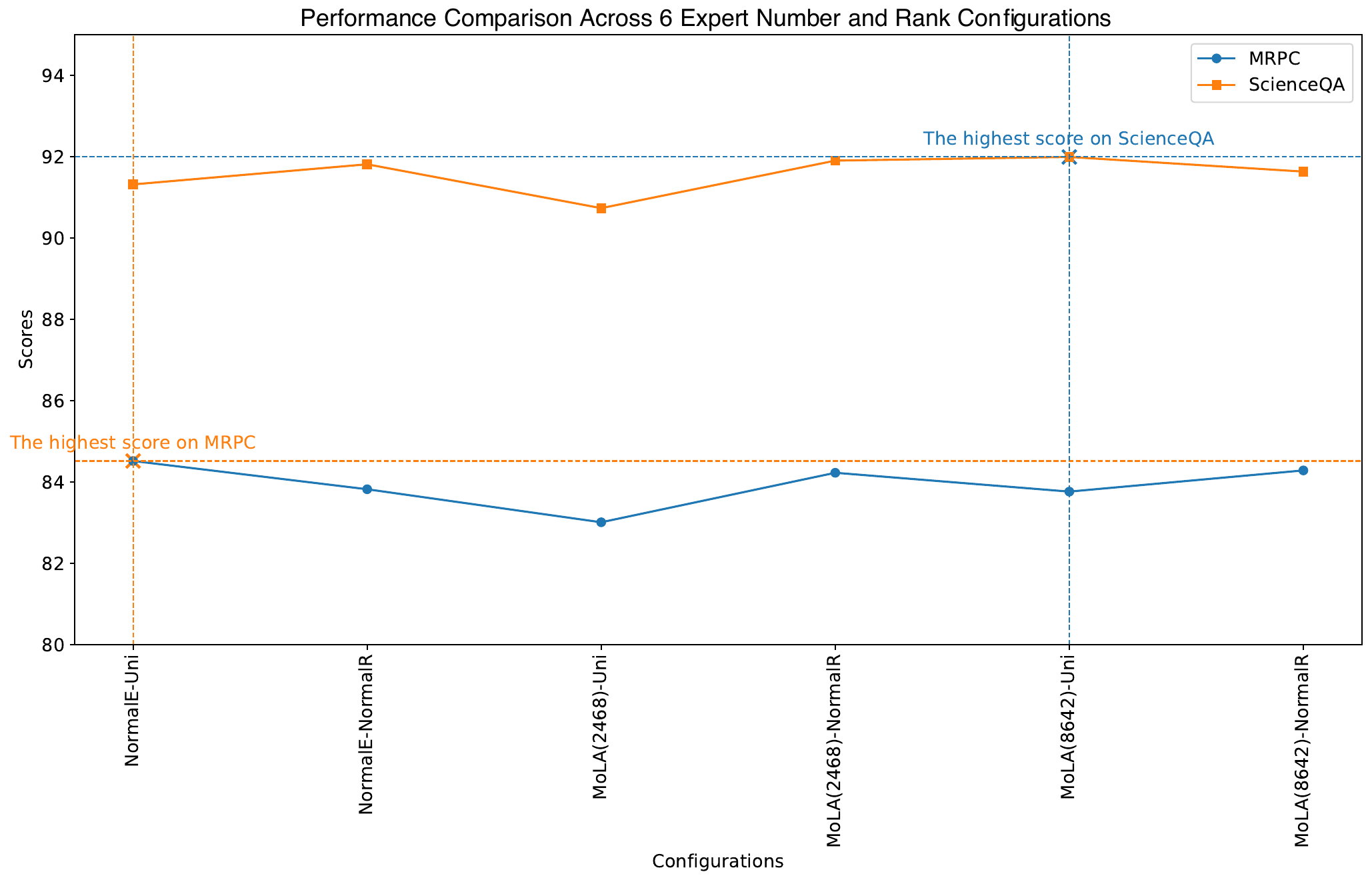}
  \caption{Performance comparison across 6 expert number and rank configurations in $\text{LLaMA-2}_\text{7B}$ on MRPC and ScienceQA. 
 }
  \label{fig:Downstream-line}
\end{figure*}

We design $6$ allocation strategies to explore whether different tasks require different expert number and rank configurations. The strategies include three options for expert number and two options for rank, resulting in $6$ combinations. 
Specifically, the expert number allocation includes two strategies from MoLA \cite{gao2024higher}, and a normal allocation: MoLA(2468), MoLA(8642), and Gaussian distribution strategy (NormalE); the rank allocation strategies are remaining uniform (Uni), and Gaussian distribution strategy (NormalR). 
We set the total rank budget for each module to 40.
NormalE selects 32 values of vertical coordinates from the standard normal distribution as selection probabilities, by uniformly sampling 32 input values within the interval $[-2\sigma, 2\sigma]$ and these values are then normalized and used to proportionally allocate the number of experts across layers, as illustrated in Fig.~\ref{fig:NormalE}. 
NormalR follows the same allocation principle as NormalE, where ranks are proportionally assigned across layers based on a normalized standard normal distribution, given a total rank budget of 40 and predefined expert number per layer.
For example, consider MoLA(2468)-Uni, 
where MoLA(2468) allocates 2 experts to
each layer for the first 8 layers, 4 experts to each layer for 9-16 layers, 6 experts to each layer for 17-24 layers, and 8 experts to each layer for the last 8 layers. In the Uni setting, if the number of experts is 4, each expert is assigned a rank of $40 \div 4 = 10$. We conduct experiments on MRPC and ScienceQA datasets using $\text{LLaMA2}_{\text{7B}}$. The results are shown in Fig.~\ref{fig:Downstream-line}. It can be observed that the performance varies across different expert number and rank configurations for the two tasks, demonstrating that different tasks require different expert numbers and ranks.

\section{Dataset }
\label{section:Dataset}
The detailed statistics of all the datasets used in our experiments are reported in Table~\ref{table6}. We source each dataset
from Huggingface Datasets 
and utilize the full dataset for our experiments.
\begin{table}[ht]
  \centering
  \setlength\tabcolsep{5pt}
\fontsize{10}{10}\selectfont 
    \begin{tabular}{l|ccc}
        \toprule[1.5pt]
        Dataset & \#Train & \#Valid & \#Test  \\
        \midrule
        CoLA & 8,551 & 1,043 & 1,063 \\
        MRPC & 3,668 & 408 & 1,725 \\
        RTE & 2,490 & 277 & 3,000 \\
        ScienceQA & 6,508 & 2,144 & 2,224 \\
        CommonsenseQA & 9,740 & 1,221 & 1,140 \\
        OpenbookQA & 5,957 & 500 & 500 \\
        MetaMathQA & 394,999 & - & - \\
        MultiArith & 420 & - & 180 \\
        SVAMP & 700 & - & 300 \\
        GSM8K & 7473 & 1319 & - \\
        \bottomrule[1.5pt]
    \end{tabular}
  \caption{\label{table6} The detailed statistics of all the datasets we used in our experiments.}
\end{table}

\section{Token Load Balance Loss}
\label{sec:BalanceLoss}
Consider a set of $N$ experts indexed by $i =1,...,N$, and a batch of $T$ tokens. The auxiliary loss is the scaled dot product of the expert usage vector $f$ and routing probability vector $P$.
\begin{equation}
  \label{eq:BAL}
  \mathcal{L}_\text{BAL} = c_B \cdot N \cdot \sum_{i = 1}^{N} f_i \cdot P_i 
\end{equation}
\noindent where $f_i$ denotes the proportion of tokens assigned to expert $i$ and $P_i$ is the fraction of the router probability allocated for expert $i$.
The auxiliary loss is scaled by $c_B = 10^{-3}$, which is large enough to encourage load balancing but small enough to avoid overshadowing the main objective.

\section{Implementation Details}
\label{sec:Implementation}
All experiments are conducted on 8× NVIDIA A800-SXM4 80GB GPUs. 
The direct fine-tuning setting aligns with AlphaLoRA \cite{qing-etal-2024-alphalora}. We perform a grid search on the number of training epochs, including 10, 15, and 20 epochs for
downstream task fine-tuning on the NLU and QA datasets. The cutoff length is
set to 256 and the batch size is 32. For the mathematical reasoning tasks, we conduct instruction tuning on the MetaMathQA
dataset \cite{yu2024metamath} for 1 epoch with cutoff length set to 512. 
In GuiLoMo, we employ two separate AdamW optimizers: one for the GuidedSelection Vector (GSVs) and one for the trainable model parameters. In all experiments, we set $e_{\text{max}} = 8$, $r_{\text{max}} = 8$ and LoRA scale parameter $\alpha$ to 16.
The optimizer for GSVs is configured with a learning rate of 3e-3, betas of (0.5, 0.999), a weight decay of 1e-3, and epsilon of 1e-8.
The optimizer for the model parameters uses a learning rate of 3e-4, betas of (0.9, 0.999), and epsilon of 1e-8. We also employ a cosine learning rate scheduler to decay the learning rate.
During the optimization process of Alg.~\ref{alg:algorithm1}, we trained for $3$ epochs with a batch size of $64$ on the NLU and QA datasets, and for $0.25$ epoch with a batch size of $32$ on the MetaMathQA dataset. 
$T$ is computed as the dataset size divided by the batch size, multiplied by the number of training epochs. 
Due to the intrinsic sparsity of GSV, we forgo the use of orthogonality regularization loss \cite{ding-etal-2023-sparse}. 

\section{The Configuration of the Baselines}
\label{section:baselines}
We set 
$\beta=2.5$ in AlphaLoRA and specify the total number of experts to be $160$.
The baselines are implemented using their open-sourced codes.   
For SoRA \cite{ding-etal-2023-sparse}, we set the maximum decayed rank to $12$, with $\lambda = 10^{-1}$, $\xi = 10^{-4}$, and $\eta_t = 10^{-1}$. 
MoLA(5) denotes using 5 experts per layer, while AlphaLoRA employs 160 in total. Under the Uniform(8) setting, each expert is assigned a rank of 8. 
When adopting the SoRA strategy for rank allocation, the maximum decayed rank is set to 12. 
\section{Prompt Templates for Fine-tuning}
We use the Alpaca prompt template for instruction tuning on three question answering datasets (ScienceQA, CommonsenseQA, and OpenbookQA):

\begin{tcolorbox}
\vspace{-0.2cm}
\begin{verbatim}
Below is an instruction that describes a task, paired with an input that provides 
further context. Write a response that appropriately completes the request.
### Instruction:
{instruction}
### Input:
{input}
### Response:
\end{verbatim}
\vspace{-0.2cm}
\end{tcolorbox}

\section{$\mathcal{H}$ operation}
\label{section:gradientg}
We adopt the sensitivity \cite{zhang2023adaptive,wang2020picking} without the norm to represent the discriminative score of the currently selected configuration:  
\begin{equation}
    \label{eq:S}
    \hat{S}(\phi) = \phi \nabla_{\phi}\mathcal{L} 
\end{equation}
where $\phi$ is any trainable parameter. 
Based on the above, the output of $\mathcal{H}(\phi)$ is such that at the position $n^{\star}$ (the index of the maximum value in 
$\phi$), its value equals $\sum_{i=1}^{n^{\star}} \hat{S}(\phi)$, while all other positions are zero. 

\section{Training Cost}
We present GPU memory (GPU memory (GuiLoMo)) of LLaMA-2-7B for Stage 1 (GSV-based bilevel optimization). For comparison, we also recorded GPU memory (GPU memory (LoRA-MoE)) when directly applying LoRA-MoE without our GuiLoMo strategy (i.e., without GSV-based bilevel optimization). As shown in Table~\ref{gpumemory}, the application of our GuiLoMo strategy, which includes GSV-based bilevel optimization, results in only a minimal increase in GPU memory usage.

\begin{table}[ht]
  \centering
  \setlength\tabcolsep{3pt} 
  \fontsize{9}{9}\selectfont 
  \begin{minipage}{0.45\textwidth}
    \centering
    \begin{tabular}{l|cc}
        \toprule[1.5pt]
         & ScienceQA & OpenBookQA \\
        \midrule
        GPU memory (GuiLoMo) & 119.42GB & 88.32GB \\
        GPU memory (LoRA-MoE) & 117.21GB & 87.10GB \\
        \bottomrule[1.5pt]
    \end{tabular}
    \caption{\label{gpumemory}GPU memory comparison with and without GuiLoMo strategy.}
  \end{minipage} \hspace{0.5cm} 
  \begin{minipage}{0.45\textwidth}
    \centering
    \begin{tabular}{l|cc}
        \toprule[1.5pt]
        $(e_{\text{max}}, r_{\text{max}})$ & (12,12) & (16,16) \\
        \midrule
        GuiLoMo & 93.21 & 93.30 \\
        AlphaLoRA + SoRA & 92.67 & 92.85 \\
        MoLA-Uniform & 92.13 & 92.31 \\
        \bottomrule[1.5pt]
    \end{tabular}
    \caption{\label{ermax}Performance comparison of GuiLoMo with other models under different $(e_{\text{max}}, r_{\text{max}})$ settings.}
  \end{minipage}
\end{table}

\section{Sensitivity to Predefined Maximum Experts and Ranks}
We evaluate GuiLoMo in two settings, i.e., $e_{max}=r_{max}=12$, and $e_{max}=r_{max}=16$, as for predefined maximum expert numbers and ranks. As shown in Table~\ref{ermax}, under the same number of trainable parameters, GuiLoMo consistently outperforms other strong baselines, demonstrating its superior allocation capability and optimization efficiency.

\onecolumn
\section{Allocation Details}
\label{section:Parameter_other}
\begin{figure*}[h]
  \centering
  \subfigure[Total number of allocated experts for sub-modules (MHA and FFN) across different layer ranges in $\text{LLaMA-2}_\text{7B}$ on MetaMathQA.]{
    \includegraphics[width=0.48\textwidth]{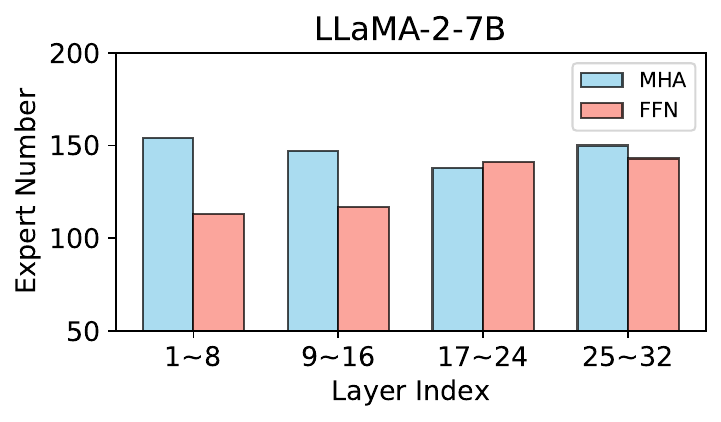}
    \label{fig:LLaMA-2-7B-expert}
  }
  \hfill
  \subfigure[Total rank of sub-modules (MHA and FFN) across different layer ranges in $\text{LLaMA-2}_{\text{7B}}$ on MetaMathQA.]{
    \includegraphics[width=0.48\textwidth]{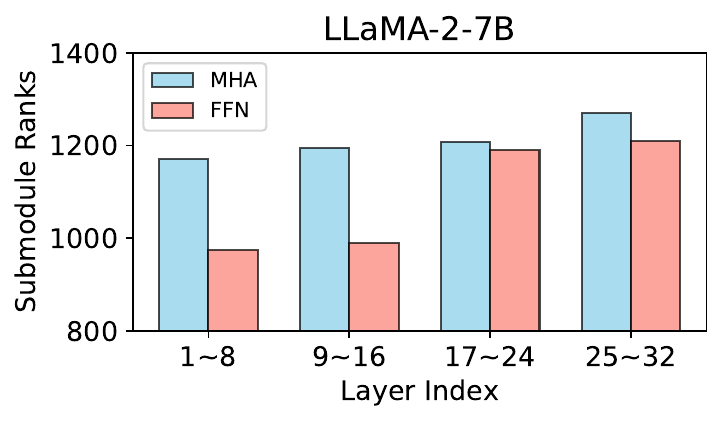}
    \label{fig:LLaMA-2-7B-total}
  }
  
\end{figure*}

\begin{figure*}[h]
  \centering
  \subfigure[Total number of allocated experts for sub-modules (MHA and FFN) across different layer ranges in $\text{Mistral-v0.1}_{\text{7B}}$ on COLA.]{
    \includegraphics[width=0.48\textwidth]{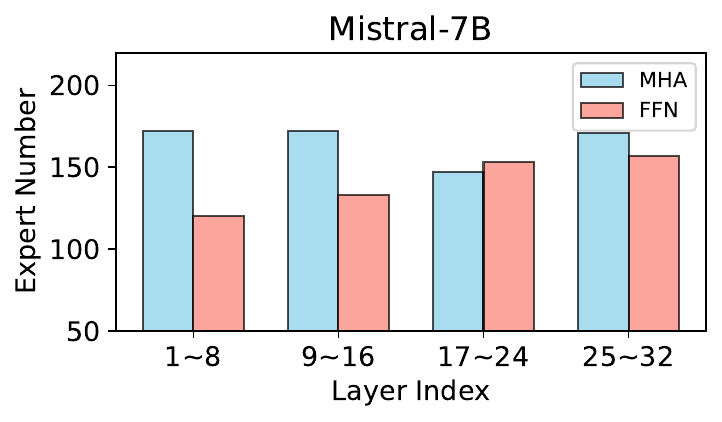}
    \label{fig:Mistral-7B-expert}
  }
  \hfill
  \subfigure[Total rank of sub-modules (MHA and FFN) across different layer ranges in $\text{Mistral-v0.1}_{\text{7B}}$ on COLA.]{
    \includegraphics[width=0.48\textwidth]{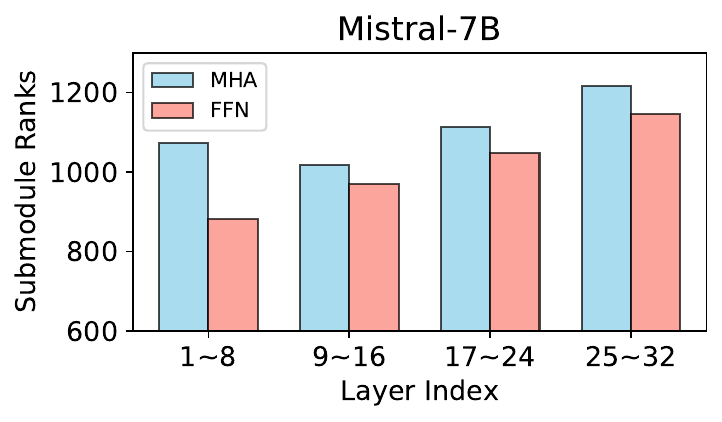}
    \label{fig:Mistral-7B-total}
  }

\end{figure*}

The total expert number and total rank of sub-modules (MHA and FFN) for $\text{LLaMA-2}_\text{7B}$ and $\text{Mistral-v0.1}_{\text{7B}}$ on MetaMathQA and COLA under different layer ranges are illustrated in Figs.~\ref{fig:LLaMA-2-7B-expert}, \ref{fig:LLaMA-2-7B-total} and \ref{fig:Mistral-7B-expert}, \ref{fig:Mistral-7B-total}, respectively.

\section{Detailed Results of Ablation Studies}
\begin{table*}[h]
  \centering
  \setlength\tabcolsep{4pt}
\fontsize{8.5}{9}\selectfont 
      \begin{tabular}{l|cccccc|c}
            \toprule
            \toprule
            Strategy & MRPC & COLA & SciQA & ComQA & GSM8K & MultiArith & Avg. \\
            \midrule
            MoLA(5)-Uniform(8) & 84.17 & 86.19 & 92.08 & 77.55 & 49.50 & 87.00 & 79.42 \\
            \midrule 
            GuiLoMo  & \textbf{85.80} & \textbf{87.25} & \textbf{92.99} & 78.46 & \textbf{53.07} & \textbf{93.67} & \textbf{81.87} \\
            w/o adaptive expert allocation & 84.99 & 86.86 & 92.13 & \textbf{78.54} & 52.16 & 89.17 & 80.64 \\
            w/o varying rank & \textbf{85.80} & 86.77 & 92.36 & 78.13 & 51.10 & 91.67 & 80.97 \\
            \bottomrule
            \bottomrule
        \end{tabular}
      \caption{\label{detailed_ablation}
        The detailed results of ablation studies on GuiLoMo across from six benchmark (MRPC, COLA, SciQA, ComQA, GSM8K, MultiArith).
      }
\end{table*}

\end{document}